\documentclass[format=acmsmall, review=false, screen=true,nonacm]{acmart}

\usepackage{amsmath}
\usepackage{booktabs}
\usepackage{graphicx}
\graphicspath{{Figures}}
\DeclareGraphicsExtensions{.pdf,.png}
\usepackage{bigstrut}
\usepackage{multirow}
\usepackage{multicol}
\usepackage{enumitem}
\usepackage{algorithm, setspace}
\usepackage{algpseudocode}
\usepackage{csquotes}
\usepackage{balance}
\usepackage{ulem} 

\newcommand{\p}[1]{\smallskip \noindent \textbf{{#1}.}}
\newcommand{\ip}[1]{\smallskip \noindent \textit{{#1}.}}
\newcommand{\eq}[1]{Equation~(\ref{eq:#1})}
\newcommand{\fig}[1]{Figure~\ref{fig:#1}}

\begin{document}

\title{Counterfactual Behavior Cloning: Offline Imitation Learning from Imperfect Human Demonstrations}


\author{Shahabedin Sagheb}
\orcid{0000-0003-3785-0319}
\email{shahab@vt.edu}

\author{Dylan P. Losey}
\orcid{0000-0002-8787-5293}
\email{losey@vt.edu}

\affiliation{
  \institution{Virginia Tech}
  \department{Department of Mechanical Engineering}
  \streetaddress{635 Prices Fork Rd}
  \city{Blacksburg}
  \state{VA}
  \postcode{24060}
  \country{USA}
  }

\thanks{This work was supported in part by NSF Grant \#2337884.}


\begin{abstract}

Learning from humans is challenging because people are imperfect teachers.
When everyday humans show the robot a new task they want it to perform, humans inevitably make errors (e.g., inputting noisy actions) and provide suboptimal examples (e.g., overshooting the goal).
\textcolor{black}{Existing methods often learn by matching some or all of the human's behavior --- but this approach is fundamentally limited because the demonstrations themselves are imperfect.}
In this work we advance offline imitation learning by enabling robots to extrapolate \textcolor{black}{across nearby actions}, instead of only considering what the human actually showed.
We achieve this by hypothesizing that all of the human's demonstrations are trying to convey \textcolor{black}{an underlying} policy, while the noise and sub-optimality within their behaviors obfuscates the data and introduces unintentional complexity.
To recover the underlying policy and learn what the human teacher meant, we introduce \textit{Counter-BC}, a generalized version of behavior cloning.
Counter-BC expands the given dataset to include actions close to behaviors the human demonstrated (i.e., counterfactual actions that the human teacher could have intended, but did not actually show).
During training Counter-BC autonomously modifies the human's demonstrations within this expanded region to reach a simplified policy that explains the underlying trends in the human's dataset.
Theoretically, we prove that Counter-BC can extract \textcolor{black}{a simple and similar-to-demonstration policy} from imperfect data, multiple users, and teachers of varying skill levels.
Empirically, we compare Counter-BC to state-of-the-art alternatives in simulated and real-world settings with noisy demonstrations, standardized datasets, and real human teachers.
Overall, we find that \textcolor{black}{trying to extrapolate what the human teacher meant by considering nearby actions} --- as opposed to rigorously following what actions the human actually demonstrated --- can lead to more proficient learning from humans.
See videos of our work here: \url{https://youtu.be/XaeOZWhTt68}

\end{abstract}

%
%




\maketitle

\section{Introduction} \label{sec:intro}

\begin{figure*}[t]
	\begin{center}
        \includegraphics[width=\linewidth]{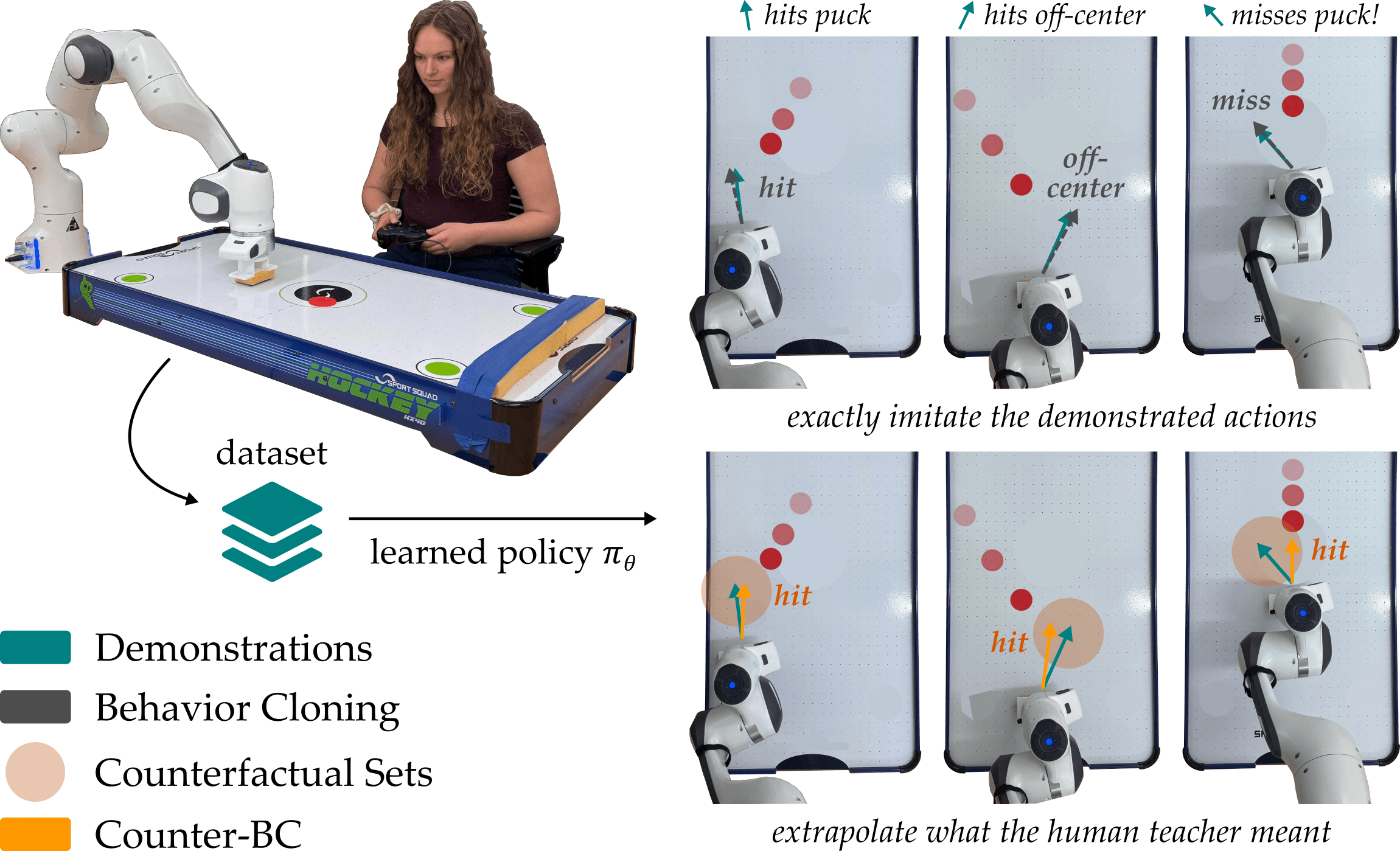}
        \caption{Learning from imperfect human demonstrations. (Left) The human teaches a robot arm to play air hockey by showing examples of hitting the puck. (Right) The robot learns a control policy based on the human's data. \textcolor{black}{Imitation learning approaches often try to mimic (at least a subset) of the actions demonstrated by the human} --- but this approach is fundamentally limited when human teachers make mistakes (like missing the puck). Counter-BC tries to imitate what the human \textit{meant}, not necessarily what the human \textit{showed}. More specifically, Counter-BC can modify the human's demonstrations within counterfactual sets in order to extrapolate underlying behaviors across the entire dataset. This leads to robots that learn consistent explanations of the human's demonstrations, e.g., learning to hit the puck at every state.}
        \label{fig:front}
	\end{center}
        \vspace{-1em}
\end{figure*}

Robots can learn new tasks by imitating human examples.
Consider the system in \fig{front}: by watching how a human plays air hockey, this robot arm should learn how to block and hit the puck.
But one fundamental limitation when learning from humans is that people are imperfect teachers \cite{sena2020quantifying, losey2022physical, chernova2022robot}.
Human demonstrations of the task will inevitably contain mistakes, noise, and suboptimal actions, particularly in real-world settings with inexperienced human operators.
For instance, when teaching the robot how to play hockey the human might accidentally miss the puck, press the wrong buttons on the joystick, or take convoluted paths towards the target.
\textcolor{black}{These sorts of mistakes can arise for multiple reasons: the task may be physically demanding or the user may not be paying sufficient attention, the teleoperation interface may introduce input delays or mapping difficulties, or the data may be collected from heterogeneous users whose skill levels and intentions differ.}
Imperfect demonstrations are a problem because --- when robots mimic suboptimal examples --- they learn to perform the task incorrectly (e.g., only hitting the puck occasionally).
\textcolor{black}{Moreover, poor demonstrations often cannot simply be discarded and re-recorded.
At scale, this would require a human expert to review thousands of demonstrations from many users, and in many deployment scenarios (e.g., data collected passively or from remote users) re-recording may not be feasible.}

To advance learning from humans we need robots that \textcolor{black}{can extrapolate} what the user actually \textit{meant}, not just what the human \textit{showed}.
This is inherently challenging because the human is the teacher: the robot infers its task based on behaviors the human demonstrates.
\textcolor{black}{Humans can make outright errors (e.g., pressing the wrong joystick button) as well as suboptimal demonstrations (e.g., taking a longer path to the goal).}
If we acknowledge that human teachers make these types of mistakes, then how should the robot reason over their potentially incorrect demonstrations in order to extract the intended task?
Recent research attempts to resolve this question by introducing additional human guidance \cite{wang2023improving, wang2023identifying, yang2021trail, kuhar2023learning, xu2022discriminator, zolna2020offline, kim2021demodice, li2023imitation, yue2024leverage, ghosh2024robust} and by iteratively testing policies in the environment \cite{brown2020better, chen2021learning, tangkaratt2020variational, zhang2021confidence, wang2021learning, wu2019imitation, cao2024limited, grigsby2021closer, kumar2022should}.
Within these prior works the robot asks a human expert to label the quality of some or all of their demonstrations \cite{wang2023improving, kuhar2023learning}.
If the robot knows which trajectories are proficient and which contain errors, then it can bias its learning to mimic the high-quality examples while ignoring low-quality behaviors.
Unfortunately, asking people to watch and label demonstrations is time consuming and subjective \cite{wang2023improving, kuhar2023learning} --- particularly for large-scale datasets that contain thousands of examples collected across diverse scenarios \cite{black2024pi_0, khazatsky2024droid, team2024octo}.
This brings us back to the fundamental problem: given only a dataset of human examples, how should the robot extract a control policy to autonomously complete the desired task?

In this work we propose a shift in perspective when learning from imperfect humans.
Instead of trying to determine how good or bad each individual human demonstration is, we recognize that \textcolor{black}{(in many scenarios)} the human teacher has an underlying policy that all of their examples are trying to convey.
The human's imperfect teaching \textit{obfuscates} this policy --- i.e., makes it seem more complex or variable than it really is --- by unintentionally adding noise and suboptimal actions into the dataset.
To recover what the human actually meant, we accordingly hypothesize that:
\begin{center}
\textit{The robot learner should adjust the human teacher's demonstrations \\ to reach a policy that simply and clearly explains their underlying trends.}
\end{center}
Applying this hypothesis we introduce \textit{Counter-BC}, a variant of behavior cloning that learns from noisy and imperfect human teachers.
Our offline imitation learning approach does not require additional human labels or knowledge about the environment.
Instead, Counter-BC automatically expands the demonstrations to consider nearby actions.
These \textit{counterfactual actions} capture the behaviors an imperfect human teacher might have been trying to show the robot.
Returning to our hockey example in \fig{front}, if the human inputs a motion that hits the puck off-center, the counterfactuals could include all actions within a radius $\Delta$ of that motion (including actions that miss the puck or hit the puck's center).
Equipped with counterfactual actions, the robot can now modify the human's demonstrations (i.e., hypothesize that the human meant to show slightly different actions) in order to reach a simple and consistent explanation of the given dataset.
\textcolor{black}{This can result in control policies that align with the underlying behaviors the human teacher intended to convey, instead of trying to match any of the actions the human actually demonstrated.}
For instance: a robot arm trained with Counter-BC learns to hit the puck at each iteration, even though our given demonstrations occasionally missed that puck.

Overall, this work is a step towards robots that learn proficient policies from datasets provided by everyday human teachers. 
We make the following contributions:

\p{Generalizing Behavior Cloning}
We analyze the loss function used to train the robot's policy in standard behavior cloning algorithms, and show why that loss is ill-posed when learning from imperfect human data.
To address this issue, we generalize the loss to include both counterfactual actions that expand the human's demonstrations and a \textcolor{black}{probabilistic} classifier that determines which counterfactual actions the robot learner should emulate.
State-of-the-art approaches --- including standard behavior cloning --- can be viewed as simplified instances of this more general loss.

\p{Deriving Counter-BC}
Given the generalized loss function, we next derive our Counter-BC algorithm by selecting the counterfactuals and classifier.
Our choices here cause the robot to learn a control policy which simultaneously does two things: i) the learned policy minimizes entropy to confidently output actions at each state, while ii) constraining those actions to remain close to the behaviors demonstrated by the imperfect human.
We provide implementation details and public code for Counter-BC, and highlight that this offline imitation learning algorithm does not require any additional information beyond the human's demonstrations.

\p{Finding Simple Explanations}
A key feature of Counter-BC is the size of the counterfactual set.
Designers can tune this size along a continuous spectrum by adjusting its radius $\Delta$.
We theoretically and empirically prove that increasing $\Delta$ (i.e., making the counterfactual sets larger) causes the robot to learn simpler policies, but these policies may increasingly diverge from the behaviors the human provided.
When learning from expert users, we decrease $\Delta$ to closely mimic the demonstrated actions; when learning from inexperienced users we increase $\Delta$ to extrapolate underlying patterns.

\p{Testing with Synthetic Demonstrations}
We compare Counter-BC to state-of-the-art baselines across multiple environments.
Within these tests we first collect synthetic human demonstrations with controlled noise distributions (e.g., Gaussian, uniform).
We then explore how effectively Counter-BC and the baselines learn the desired task as we vary the noise within the human's demonstrations.
In general, we find that Counter-BC outperforms existing methods regardless of what type of noise or suboptimal actions the synthetic human provides.

\p{Learning from Real Humans}
We conclude by learning from actual human data across simulated environments, standardized datasets, and an air hockey experiment.
We again compare Counter-BC to existing offline imitation learning methods, and study how proficiently the robot learns the desired task based on examples from $N=20$ human demonstrators.
{Our results indicate that robots which try to extrapolate what the human teacher meant by considering counterfactual actions (as opposed to copying or ignoring what the human actually did) are able to learn more effective policies from real human data.}

{Counter-BC is most applicable to scenarios where: (1) the robot learns from a fixed, offline dataset that cannot be re-collected or curated; (2) the task requires a continuous control policy with a relatively low-dimensional action space; and (3) the human teachers share a common intent (i.e., they are all trying to teach the same task, even if their skill levels differ). Concrete examples include assistive robotics applications where caregivers of varying skill demonstrate desired motions, manufacturing settings where workers teach robot manipulators new assembly steps, and imitation learning pipelines that aggregate crowd-sourced demonstrations.}
\section{Related Work} \label{sec:related}

Learning from noisy and imperfect human demonstrations is a core challenge for imitation learning \cite{zare2024survey, belkhale2023data, mandlekar2021matters}.
Humans inevitably make errors when they show robots how to perform tasks --- if robots naively mimic these mistakes, they will learn to perform the tasks incorrectly.
The emergence of large-scale datasets for robot learning has made this issue increasingly important.
To collect vast amounts of data, systems such as \cite{black2024pi_0, khazatsky2024droid, team2024octo} rely on multiple human teachers of varying skill and proficiency.
Given a heterogeneous dataset, how should the robot extract the correct behavior?

\p{Improving Human Teaching}
One approach is to help the human become a better teacher.
By giving the human offline guidance before they demonstrate the task, or real-time feedback during their demonstration, the robot may improve the quality of the human's examples \cite{habibian2024survey, sena2020quantifying, chernova2022robot}.
For instance, the robot can indicate when the human's current trajectory deviates from previous behaviors.
Related works have accordingly leveraged augmented reality \cite{sakr2023can}, haptic interfaces \cite{valdivia2023wrapping}, physical markers \cite{sanchez2024recon}, or computer screens \cite{gandhi2023eliciting} to assist the human teacher.
{For example, in robot-centric learning from demonstration \cite{laskey2017comparing, knox2008tamer, ross2011reduction, schrum2022mind} the robot takes an active role during data collection by querying the human for additional demonstrations or feedback at states it finds uncertain.}
Once demonstrations are collected, the robot can also post-process the resulting data to filter out poor examples and human errors \cite{hejna2025robot, chen2025curating}.
{Overall, these works are complementary but orthogonal to our efforts: instead of assisting the human during online data collection, we will advance how the robot learns from a fixed, offline, human-provided dataset.}

A variety of learning algorithms have already been proposed for imperfect human demonstrations.
Below we divide this related research into three groups: methods that rely on environment interactions, methods that rely on rankings, and methods that only consider the offline dataset. 

\p{Leveraging Environment Interactions}
The first way in which robots can learn from noisy demonstrations is by extracting a reward \cite{brown2020better, chen2021learning, tangkaratt2020variational} or discriminator \cite{zhang2021confidence, wang2021learning, wu2019imitation, cao2024limited} from those demonstrations.
The robot then interacts with the environment online (i.e., the robot arm attempts to complete the task) using reinforcement learning algorithms \cite{brown2020better, chen2021learning, tangkaratt2020variational} or adversarial networks \cite{zhang2021confidence, wang2021learning, wu2019imitation, cao2024limited}.
Through these interactions the robot finds policies which maximize the reward, or which the discriminator cannot distinguish from expert human examples.
One advantage of this paradigm is that --- given ranked demonstrations --- the robot can learn behaviors that are better than the best human demonstrations \cite{brown2020better, chen2021learning}.
But the downside is that this approach is \textit{online}, and requires access to the environment dynamics through simulations or real-world rollouts.
For practical implementation, we instead focus on imitation learning algorithms which are purely \textit{offline}, and do not require any environment interactions during training.

\p{Leveraging Supervised Labels}
Some research achieves offline imitation learning from imperfect demonstrations by labeling (or scoring) the dataset.
These labels reflect the relative quality of an example: e.g., a trajectory that goes precisely to the goal might be labeled as an ``expert'' demonstration and given a high score, while a trajectory that overshoots the goal could be labeled as ``novice'' and given a low score.
Based on this labeled dataset, the robot learns a policy offline which aligns with high-quality demonstrations while down-weighting or ignoring low-quality inputs.
The key challenge here is how to label the dataset; i.e., how to discern expert and novice demonstrations.
When the robot has access to its reward function, this process is straightforward
\cite{grigsby2021closer, kumar2022should}; trajectories with higher rewards are higher quality.
More generally, recent works focus on settings where a portion of the dataset is labeled, and the rest is unlabeled \cite{wang2023improving, wang2023identifying, yang2021trail, kuhar2023learning, xu2022discriminator, zolna2020offline, kim2021demodice, li2023imitation, yue2024leverage, ghosh2024robust, wang2023imitation, yue2024leverage}.
The exact way that the labels are collected can vary.
For instance, in \cite{wang2023improving} it is assumed that a human has marked some subset of the data as ``expert.''
Alternatively, in \cite{kuhar2023learning} the human indicates the relative quality of a few demonstrations along a continuous or discrete scale.
Regardless of how the labels are obtained, these works leverage the known labels to autonomously determine the quality of the remainder of the (unlabeled) dataset, and then train a policy to align with the high-quality examples.
Of course, the downside to this framework is that labeling the demonstrations takes time and effort; a human expert must watch multiple demonstrations to grade their relative quality.
We therefore seek a method which \textit{does not require} any additional human inputs, rankings, or labels.

\p{Leveraging Only Human Demonstrations} 
Most related to our proposed approach are methods that learn a policy given only the human's noisy and imperfect demonstrations.
These methods do not assume access to environment interactions or supervised labels; instead, they seek to extrapolate what the human meant based only on trends in the human's data.
In \cite{sasaki2020behavioral} the authors modify behavior cloning so that the robot's learned policy concentrates on the modes of the human's demonstrations.
Put another way, the robot learns to consistently mimic the behaviors which are most common in the dataset.
When these modes are correct, the robot learns the desired task --- but if the majority of the dataset is not expert behavior, \cite{sasaki2020behavioral} falls short.
Similarly, \cite{beliaev2022imitation} assumes that human demonstrations are high-quality when those demonstrations output a consistent action at a given state, and low-quality when the demonstrated actions have high variance.
We will experimentally compare our proposed algorithm to both of these state-of-the-art baselines.
In general, the key difference between our method and \cite{sasaki2020behavioral} or \cite{beliaev2022imitation} is that we enable the robot to incorporate counterfactuals (i.e., actions that the imperfect human could have intended, but did not actually demonstrate) to build a control policy that consistently explains the data.
We note that \cite{sun2023offline} also uses counterfactuals, but applies them in settings where the robot has access to expert labels.

\section{Problem Formulation} \label{sec:problem}

We consider scenarios where a robot is learning a new task from one or more human teachers.
Offline, the human teacher(s) provide examples of how the robot should complete the desired task.
Because humans are noisy and imperfect, their demonstrations will inevitably contain mistakes (i.e., suboptimal actions that they do not want the robot to imitate).
The robot's goal is to learn how to perform the task correctly, despite imperfect or suboptimal examples in the dataset.

\p{Robot} 
Let $s \in \mathcal{S}$ be the system state and let $a \in \mathcal{A}$ be the robot's action.
For instance, within our motivating example\textcolor{green!60!black}{,} $s$ contains the robot's joint position, end-effector pose, and an image of the hockey table; the action $a$ is the robot's joint velocity.
The robot's actions cause the system state to transition according its dynamics.
We do not assume access to these environment dynamics.
Put another way, when the robot arm moves to hit the hockey puck, the robot cannot explicitly model or simulate of how the puck will respond to its actions.

\p{Human}
The human has in mind a task they want the robot to learn.
To teach the robot, the human demonstrates how to complete that task.
Prior works have established multiple forms of demonstration: the human might kinesthetically guide the robot through the desired motion \cite{akgun2012keyframe}, teleoperate the robot along a trajectory \cite{losey2022learning}, or employ graphical user interfaces, augmented reality, \textcolor{black}{motion capture, passive observation}, or natural language inputs to indicate the correct action at various states \cite{chen2024arcap}.
Regardless of how they are collected, the human's demonstrations result in a dataset $\mathcal{D}$ of examples.
Within this dataset the human teacher labels each system state $s$ with an action $a$, such that $\mathcal{D} = \{(s_1, a_1), \ldots, (s_n, a_n)\}$ is a collection of $n$ state-action pairs.
In our experiments we will focus on \textit{offline datasets} (i.e., datasets collected prior to robot learning), but our approach also applies to settings where the human shows new state-action pairs online as the robot attempts to complete its task \cite{spencer2022expert}.

It is inevitable that the human will make mistakes when giving demonstrations.
These mistakes may occur because the task is hard to perform, because the human is distracted or moving quickly, or because it is challenging to perfectly orchestrate the motion of a dexterous robot arm \cite{losey2022physical, bobu2024aligning, sena2020quantifying, chernova2022robot}.
To formulate teaching errors we define $a^*$ as the \textit{ideal action} the human meant to show the robot.
If the human acted perfectly, they would have provided an ideal dataset $\mathcal{D}^* = \{(s_1, a_1^*), \ldots, (s_n, a_n^*)\}$.
In practice, however, the noisy human actually demonstrates suboptimal actions $a$:
\begin{equation} \label{eq:P1}
    a = a^* + \epsilon, \quad \epsilon \sim P(\cdot \mid s)
\end{equation}
where $\epsilon$ is the error between the ideal action $a^*$ and the actual action $a$.
At each state in the collected dataset $\mathcal{D}$ \textcolor{green!60!black}{,} this action error is sampled from some unknown distribution $P(\epsilon \mid s)$.

\p{Policy}
Despite the noise inherent in the human's examples, the robot's objective is to learn a \textit{control policy} that autonomously completes the demonstrated task. 
This control policy is a mapping from observed states $s$ to robot actions $a$:
\begin{equation} \label{eq:P2}
    a \sim \pi_\theta(\cdot \mid s)
\end{equation}
We instantiate the control policy $\pi$ as a model (e.g., a neural network) parameterized by weights $\theta \in \Theta$.
In practice, there are three different control policies that our formulation must consider: i) the ideal policy $\pi^*$ that the human is trying to convey to the robot, ii) the noisy and imperfect policy $\pi$ that the human actually follows when providing demonstrations, and iii) the robot's learned policy $\pi_\theta$.
Ideal actions $a^*$ and the ideal dataset $\mathcal{D}^*$ are sampled from the ideal policy $\pi^*$, while the demonstrated actions $a = a^* + \epsilon$ and the collected dataset $\mathcal{D}$ are sampled from the imperfect policy $\pi$.
We emphasize that the robot does not know either $\pi^*$ or $\pi$.
Instead, based on the dataset $\mathcal{D}$ induced by $\pi$, the robot seeks to learn a control policy $\pi_\theta$ that matches the ideal policy $\pi^*$.

\p{Behavior Cloning}
Recent works have found that methods founded on behavior cloning are a promising way to learn $\pi_\theta$ from suboptimal demonstrations \cite{mandlekar2021matters, florence2022implicit, sasaki2020behavioral, mehta2025stable}.
To better understand these baselines --- and set the stage for our own approach --- we here derive the loss function used to train the robot's policy in behavior cloning algorithms.
Intuitively, the robot seeks to learn $\theta$ such that its control policy $\pi_\theta$ matches the expert policy $\pi^*$.
We can measure the distance between distribution $\pi_\theta$ and distribution $\pi^*$ using Kullback–Leibler (KL) divergence. 
Applying KL divergence to conditional probabilities, we reach:
\begin{align} \label{eq:P3}
    D_{KL}(\pi^*, \pi_\theta) &= C-\int_{s \in \mathcal{S}} \int_{a \in \mathcal{A}}  \Big[\rho^*(s)\pi^*(a \mid s) \log{\pi_\theta(a \mid s)} \Big] \\
    &=C - \mathbb{E}_{s \sim \rho^*(\cdot), a^* \sim \pi^*(\cdot \mid s)} \Big[\log{\pi_\theta(a^* \mid s)} \Big] \label{eq:P4}
\end{align}
where $C$ is a constant term that does not depend on $\theta$, and $\rho^*(s)$ is the distribution over states induced by the expert policy $\pi^*(a \mid s)$.
Accordingly, to reduce the error between $\pi_\theta$ and $\pi^*$ the robot should update $\theta$ to minimize the right side of \eq{P4}.
This leads to the loss function:
\begin{equation} \label{eq:P5}
    \mathcal{L}(\theta) = \mathbb{E}_{s \sim \rho^*(\cdot), a^* \sim \pi^*(\cdot \mid s)} \Big[-\log{\pi_\theta(a^* \mid s)}\Big]
\end{equation}
Approximating this expectation by sampling from the expert policy, \eq{P5} becomes:
\begin{equation} \label{eq:P6}
    \mathcal{L}(\theta) \propto \sum_{(s, a^*) \in \mathcal{D}^*} \Big[-\log{\pi_\theta(a^* \mid s)}\Big]
\end{equation}
which forms the standard loss function for behavior cloning algorithms \cite{ke2021imitation, sasaki2020behavioral, belkhale2023data}\textcolor{green!60!black}{.}
Training the model weights $\theta$ to minimize the loss in \eq{P6} corresponds to learning a control policy $\pi_\theta$ that matches the expert policy $\pi^*$ across states in the dataset.

Unfortunately, this standard behavior cloning loss has a significant issue when applied to noisy and imperfect data.
Based on the derivation in \eq{P6}, to learn the intended policy the robot should reason over the ideal dataset $\mathcal{D}^*$.
But in reality the robot does not have access to $\mathcal{D}^*$. 
Instead, humans provide $\mathcal{D}$ --- a dataset with noisy, imperfect, and suboptimal examples.
One naive solution is just to replace $\mathcal{D}^*$ with $\mathcal{D}$ in \eq{P6} and then train the robot's policy (we test this baseline in our experiments).
However, doing so means that our robot is being trained to imitate incorrect human examples as opposed to the desired behaviors.
We must therefore develop an approach that modifies \eq{P6} to extrapolate from the noisy human demonstrations in dataset $\mathcal{D}$.

\section{Counterfactual Behavior Cloning} \label{sec:method}

In this section we present counterfactual behavior cloning (Counter-BC), a modification of behavior cloning designed to learn from noisy and imperfect human demonstrations.
\textcolor{black}{Our core hypothesis is that human teachers have in mind an underlying control policy they are trying to convey, and the noise in their demonstrations obfuscates this control policy by introducing additional variation and complexity.}
As such, robot learners should reason over the human's demonstrations --- and even modify those demonstrations --- in order to reach a \textit{simple explanation} of the given data.
Counter-BC achieves this by expanding the demonstrations to include counterfactual actions: i.e., actions that the suboptimal human did not actually demonstrate, but may have intended to show the system.
The robot learns a policy that minimizes entropy over these counterfactual actions; this policy is a \textit{simple explanation} because the learned function is confident about what action to take at each state in the dataset (preventing the robot from overfitting to the variability and complexity introduced by the human's noisy behavior).
In Section~\ref{sec:M1}\textcolor{green!60!black}{,} we expand the standard behavior cloning loss to now account for counterfactual actions, and explain how this generalizes methods from prior work.
Next, in Section~\ref{sec:M2}\textcolor{green!60!black}{,} we derive the key terms of the new loss function, and show how the resulting policy selects counterfactuals that lead to minimal entropy.
Finally, in Section~\ref{sec:M3}\textcolor{green!60!black}{,} we outline the Counter-BC algorithm and practical implementation details.

\subsection{Generalizing the Loss with Counterfactuals} \label{sec:M1}

Human teachers inevitably provide imperfect and suboptimal demonstrations.
In \eq{P1}\textcolor{green!60!black}{,} we formulated the impact of those demonstrations: instead of showing actions $a^*$ (sampled from the ideal policy $\pi^*$), everyday users input actions $a$ (sampled from their suboptimal policy $\pi$).
Put another way, at each $(s, a)$ pair in the dataset $\mathcal{D}$\textcolor{green!60!black}{,} the human may have actually meant to show some other behavior.
Our proposed method captures these alternative actions through \textit{counterfactual sets}.
Given a state-action pair, let $a' \in \mathcal{C}(s, a)$ be the set of counterfactual actions that the human could have intended to demonstrate.
By leveraging counterfactual sets $\mathcal{C}$\textcolor{green!60!black}{,} we can expand the standard behavior cloning loss from \eq{P6} into a more general form:
\begin{equation} \label{eq:M1}
    \mathcal{L}(\theta) \propto \sum_{(s, a) \in \mathcal{D}}\sum_{a' \in \mathcal{C}(s, a)} \Big[-R(s, a') \cdot \log{\pi_\theta(a' \mid s)} \Big]
\end{equation}
Here $\mathcal{C}$ iterates through the actions the human might have wanted to show, and $R(s, a) \in [0, 1]$ classifies these hypothetical actions along a continuous spectrum from expert ($R \rightarrow 1$) to non-expert ($R \rightarrow 0$).
\eq{M1} reduces to \eq{P6} when: i) the counterfactuals include the ideal action $a^* \in \mathcal{C}(s, a)$ at each state, and ii) $R(s, a) = 1$ for $a = a^*$, and $R(s, a) = 0$ otherwise.
Introducing counterfactual actions is a shift in the robot's perspective: instead of strictly learning to match the behaviors in $\mathcal{D}$ demonstrated by the imperfect human teacher, now the robot can learn to mimic actions in $\mathcal{C}(s, a)$ which the human never actually showed.

We highlight that the loss functions used by related works can also be viewed as simplifications of \eq{M1}.
As summarized below, we reach these simplifications by making different choices for the counterfactual sets $\mathcal{C}$ or the classifier $R$ in \eq{M1}:
\begin{itemize}
    \item \textbf{Behavior cloning} \cite{ke2021imitation}: $\mathcal{C}(s, a) = \{a\}$, $R(s, a) = 1$ for all states.
    \item \textbf{Advantage-filtered behavior cloning} \cite{grigsby2021closer, kumar2022should}: $\mathcal{C}(s, a) = \{a\}$, $R(s, a)$ is based on the advantage function or value associated with each state-action pair.
    \item \textbf{Discriminator-weighted behavior cloning} \cite{xu2022discriminator}: $\mathcal{C}(s, a) = \{a\}$, $R(s, a)$ is a discriminator that assigns $R(s, a) \rightarrow 1$ to expert behavior and $R(s, a) \rightarrow 0$ to novice behavior.
    \item \textbf{ILEED} \cite{beliaev2022imitation}: $\mathcal{C}(s, a) = \{a\}$, $R(s, a)$ is the expertise level of the current human teacher where $R(s, a) \rightarrow 1$ when the human is an expert at the current state.
    \item \textbf{BCND} \cite{sasaki2020behavioral}: $\mathcal{C}(s, a) = \{a\}$, $R(s, a) = \pi_{prev}(a \mid s)$ is the learned control policy from the previous training iteration.
\end{itemize}
One trend we observe is that --- across each of these state-of-the-art methods --- the robot always sets $\mathcal{C}(s,a) = \{a\}$.
This means that no actions besides the ones demonstrated by the human teacher can be explicitly treated as ``expert'' behavior (i.e., the robot does not reason over counterfactuals).
As we will show, using counterfactuals is a fundamental change that enables the robot to interpret and explain the human's demonstrations in ways not available when the system is constrained to only imitating ($R \rightarrow 1$) or ignoring ($R \rightarrow 0$) the behaviors provided by the human teacher.

\subsection{Learning Policies that Consistently Explain the Counterfactuals} \label{sec:M2}

Given the generalized loss in \eq{M1}, our next steps are to select the counterfactual set $\mathcal{C}$ and classifier $R$ used by our algorithm.
Our choices of $\mathcal{C}$ and $R$ should enable the robot learner to reason over the counterfactuals and reach a simple explanation of the underlying control policy.

\p{Counterfactual Set}
The counterfactual set accounts for noise and error in the human's demonstrations by considering alternative actions that are close to the human's actual behavior.
For each $(s, a) \in \mathcal{D}$, let $\mathcal{C}(s, a)$ contain all actions within a radius $\Delta$ of the demonstrated $a$:
\begin{equation} \label{eq:M2}
    \mathcal{C}(s, a) = \{a' \mid \| a - a' \| - \Delta \leq 0\}
\end{equation}
The radius $\Delta \geq 0$ in \eq{M2} is a hyperparameter specified by the designer.
\textcolor{black}{
Decreasing $\Delta$ means that the robot will have a smaller counterfactual set, i.e., the learner assumes that the ideal action $a^*$ is close to the demonstrated action $a$.
Conversely, increasing $\Delta$ results in a larger counterfactual set, i.e., the learner can treat a variety of actions as $a^*$ even if those actions are dissimilar from the human's demonstration $a$.
We empirically find that the best performing values of $\Delta$ lie between these extremes, enabling the robot to extrapolate without discarding the human's underlying policy.
Smaller values of $\Delta$ are appropriate for expert teachers (where the errors in $a$ are slight), and larger values of $\Delta$ should be used with novice teachers (where the errors in $a$ may be significant).}

\textcolor{black}{We note that the counterfactual set $\mathcal{C}(s, a)$ should be constrained to the action set $\mathcal{A}$, such that $\mathcal{C}(s, a) \subseteq \mathcal{A}$. Put another way, the robot should not consider counterfactuals that do not belong to its set of feasible actions. We here assume that the designer knows the action space.}
\textcolor{black}{We further note that the counterfactual assumption is most natural when the action space is continuous and Euclidean proximity reflects physical similarity (e.g., joint velocity or end-effector velocity commands). In settings where this assumption does not hold (for instance, when nearby values in the action space correspond to semantically distinct behaviors) designers should reduce $\Delta$ or consider a domain-specific distance metric when defining $\mathcal{C}(s, a)$. Alternatively, redesigning the action parameterization so that physically similar motions are represented as numerically nearby values can recover the benefits of Counter-BC in such settings.}

\p{\textcolor{black}{Probabilistic Classifier}}
The \textcolor{black}{probabilistic classifier} $R : \mathcal{S} \times \mathcal{A} \rightarrow [0, 1]$ determines which actions the control policy should emulate.
Given a proposed counterfactual $a'$, increasing $R(s, a')$ indicates that $a'$ is likely the intended human action.
We have already introduced a term that operates similarly to $R$: the control policy $\pi_\theta(a \mid s)$ estimates the likelihood that a given action is part of the desired policy at state $s$.
Recognizing this alignment, we define the \textcolor{black}{probabilistic classifier} as a restriction of probability distribution $\pi_\theta$ with support over the counterfactual set:
\begin{equation} \label{eq:M3}
    R(s, a') = \hat{\pi}_\theta(a' \mid s), \quad \hat{\pi}_\theta(a' \mid s) = \frac{\pi_\theta(a' \mid s)}{\sum_{z \in \mathcal{C}(s, a)} \pi_\theta(z \mid s)}
\end{equation}
Because the normalizer of $\hat{\pi}_\theta$ considers only the counterfactual set $\mathcal{C}$, by definition $\hat{\pi}_\theta$ is a probability distribution that sums to one across the counterfactuals.
Put another way, $R(s, a')$ in \eq{M3} expresses the probability that counterfactual $a'$ is really the action $a^*$ the human teacher meant to demonstrate.
For instance: in the special case where $\mathcal{C}(s,a) = \{a\}$ and the robot does not consider counterfactuals, then $R(s, a) = 1$ and the robot is confident that the demonstrated action $a$ is the intended action $a^*$.
In what follows we will prove that this choice for \textcolor{black}{probabilistic classifier} $R$ (when combined with counterfactual sets $\mathcal{C}$) biases the robot learner towards policies that provide simple explanations for the human's noisy demonstrations.

\begin{figure*}[t]
	\begin{center}
        \includegraphics[width=\linewidth]{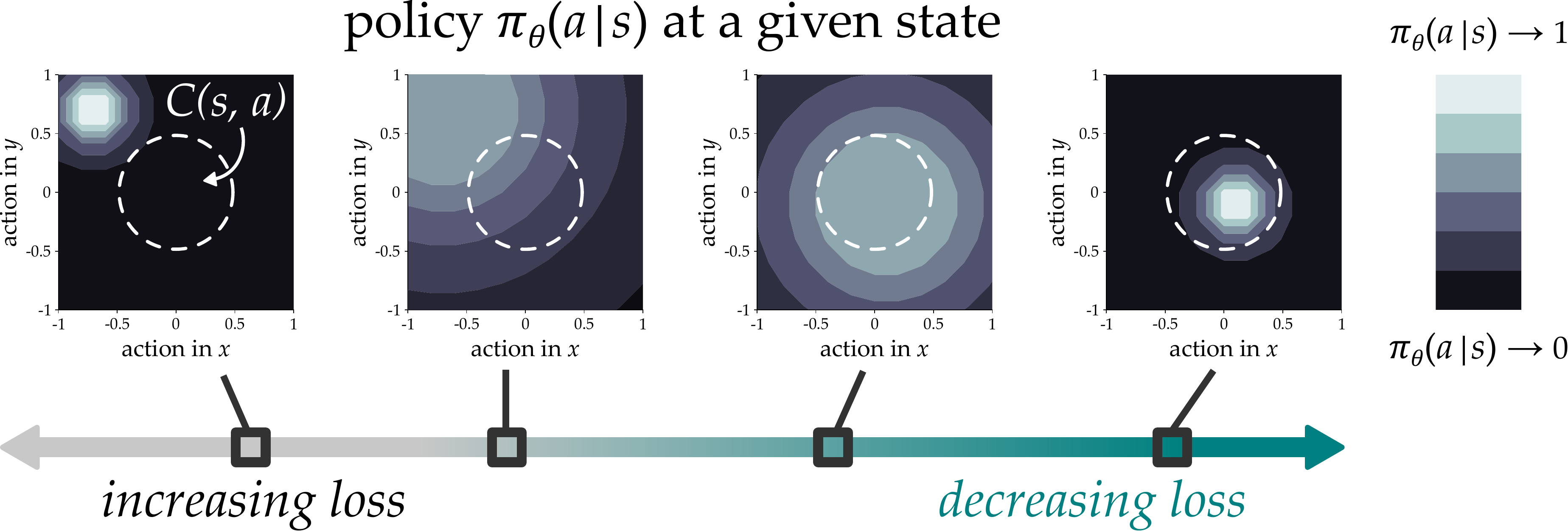}
		\caption{Visualizing the loss in \eq{M7} for a single state. Here the actions are two dimensional (i.e., in an $x$-$y$ plane), and the demonstrated human action is at the origin. The counterfactual set lies within the dashed white circle ($\Delta = 0.5$). We show the probability distribution over actions for four different policies $\pi_\theta(a \mid s)$, where lighter regions indicate that the action is more likely. (Far Left) The maximum loss occurs when the policy confidently predicts an action outside the counterfactual. (Far Right) The minimum loss occurs when policy confidently predicts an action inside the counterfactual. Note that the predicted action does not need to be the demonstrated action; specifying any action within $\mathcal{C}(s, a)$ can minimize the proposed loss.}
		\label{fig:method1}
	\end{center}
        \vspace{-1em}
\end{figure*}

\p{Counter-BC Loss}
To derive the loss function for Counter-BC we plug \eq{M2} and \eq{M3} back into \eq{M1}.
Starting with our given choices of the counterfactual set and classifier, and then manipulating the terms, we eventually reach the cross entropy between $\hat{\pi}_\theta$ and $\pi_\theta$:
\begin{align} \label{eq:M4}
    L(\theta) & \propto \sum_{(s, a) \in \mathcal{D}}\sum_{a' \in \mathcal{C}(s, a)} \Big[-\hat{\pi}_\theta(a' \mid s) \cdot \log{\pi_\theta(a' \mid s)}\Big] \\  \label{eq:M5}
    & = \sum_{(s, a) \in \mathcal{D}}\sum_{a' \in \mathcal{C}(s, a)} \Bigg[-\hat{\pi}_\theta(a' \mid s) \cdot \log{\frac{\pi_\theta(a' \mid s) \cdot \hat{\pi}_\theta(a' \mid s)}{\hat{\pi}_\theta(a' \mid s)}}\Bigg] \\  \label{eq:M6}
    & = \sum_{(s, a) \in \mathcal{D}}\sum_{a' \in \mathcal{C}(s, a)} \Bigg[-\hat{\pi}_\theta(a' \mid s) \cdot \log{\hat{\pi}_\theta(a' \mid s)} + \hat{\pi}_\theta(a' \mid s) \cdot \log{\frac{\hat{\pi}_\theta(a' \mid s)}{\pi_\theta(a' \mid s)}}\Bigg]  \\  \label{eq:M7}
    & = \sum_{(s, a) \in \mathcal{D}} H\Big(\hat{\pi}_\theta \mid s, \mathcal{C}(s, a)\Big) + D_{KL}\Big(\hat{\pi}_\theta, \pi_\theta \mid s, \mathcal{C}(s, a)\Big)
\end{align}
Here $H$ is the conditional entropy over the restricted policy's actions given state $s$ and counterfactual set $\mathcal{C}$, and $D_{KL}$ is the KL divergence between $\hat{\pi}_\theta$ and $\pi_\theta$ given state $s$ and counterfactual set $\mathcal{C}$.
In practice, minimizing \eq{M7} does two things.
The first term drives our restricted policy $\hat{\pi}_\theta$ to have low entropy over the counterfactual set.
The second term forces the robot's full policy $\pi$ to match the restricted policy $\hat{\pi}_\theta$.
Putting these terms together means that the robot will learn a policy $\pi_\theta$ that confidently outputs actions (i.e., minimizes entropy), while constraining those actions to be close to the behaviors demonstrated by the imperfect human.

To better visualize this loss function\textcolor{green!60!black}{,} we provide an example in \fig{method1}.
The loss is \textit{highest} when the policy $\pi_\theta$ confidently predicts an action outside of the counterfactual set (far left). Conversely, the loss is \textit{lowest} when $\pi_\theta$ confidently predicts an action within $\mathcal{C}(s, a)$ (far right).

\textcolor{black}{
Viewed at a single state $s$, minimizing \eq{M7} results in a robot policy $\pi_\theta(a \mid s)$ that confidently outputs an action $a \in \mathcal{C}(s, a)$.
We next consider what happens when we apply this loss function across the entire dataset $\mathcal{D}$ of state-action pairs.
At each state in this dataset, the robot can learn to output any action within the corresponding counterfactual $\mathcal{C}(s, a)$.
The radius $\Delta$ in \eq{M2} defines the size of this counterfactual set. 
When $\Delta = 0$ the counterfactual set is simply the action demonstrated by the human, and the robot tries to precisely imitate the human’s actions. 
As $\Delta$ increases, the size of the counterfactual set also increases to include a wider range of actions --- at the extreme, the counterfactual set is $\mathcal{A}$. 
So given this increasing range of options, which actions will a policy trained with \eq{M7} learn to output?
If we assume that the model $\pi_\theta$ is locally Lipschitz \cite{jordan2020exactly}, then for small changes in the state $s$, the model is inherently structured to output small changes in the action $a$.
In other words --- if the robot is free to choose any action within this counterfactual set --- the policy is likely to select an action at state $s'$ that is similar to the action at nearby state $s$.
Expanding the radius $\Delta$ therefore means that the robot learns a ``simpler'' policy, in the sense that the learned policy maintains consistent actions across nearby states.
}

\p{Intuition}
By extrapolating across the counterfactual set, our policy seeks to \textit{explain} the human's data, and can \textit{modify} the expert actions (i.e., change the selected counterfactual) in order to reach a more efficient explanation of the dataset.
Overall, the design of \eq{M7} formalizes our hypothesis that there is an underlying function behind the human's behaviors, and the robot may need to tweak the dataset in order to remove suboptimal actions and recover that intended function.

\begin{figure*}[t]
	\begin{center}
        \includegraphics[width=\linewidth]{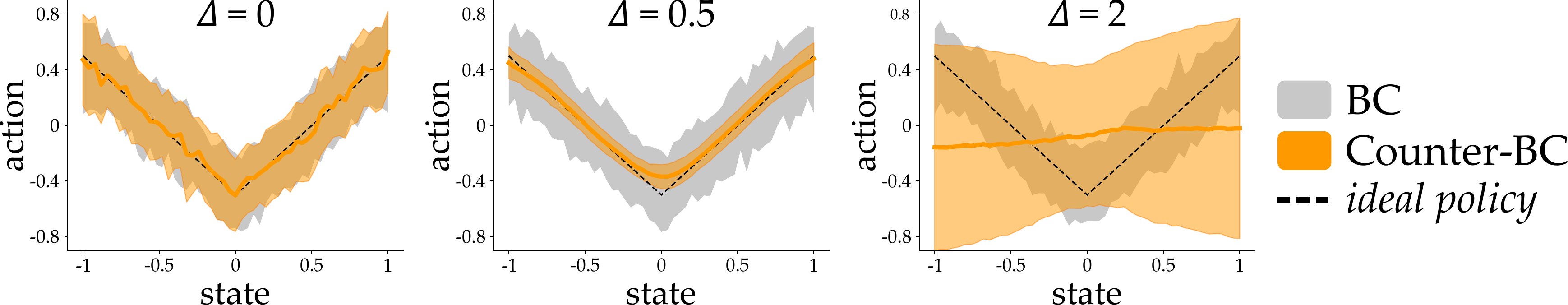}
		\caption{Recovering the underlying function with Counter-BC. In a $1$-dimensional environment the simulated human noisily demonstrates the absolute value function $a = |s| - 0.5 + \epsilon$, where $\epsilon \sim \mathcal{U}(-0.5, +0.5)$. We plot the robot's learned policy averaged across $50$ runs; shaded regions show the standard deviation.
        With behavior cloning (BC) the robot tries to precisely match the noisy data, resulting in a more complex explanation of the human's demonstrations. (From left to right) Counter-BC with different values of hyperparameter $\Delta$. Increasing $\Delta$ causes Counter-BC to recover increasingly simple explanations of the data by reasoning over larger counterfactual sets, but the resulting policies may diverge from the human's examples.}
		\label{fig:method2}
	\end{center}
        \vspace{-1em}
\end{figure*}

In \fig{method2}\textcolor{green!60!black}{,} we show an instance of using the Counter-BC loss to recover the underlying task --- here a simulated human is demonstrating the absolute value function.
Within this example we also highlight how tuning $\Delta$ affects the recovered policy.
In general: \textit{increasing $\Delta$ means the robot will recover a simpler function, but this function may increasingly diverge from the dataset}.
When hyperparameter $\Delta \rightarrow 0$ then $\mathcal{C}(s, a) = \{a\}$ and minimizing \eq{M7} is equivalent to standard behavior cloning.
Hence, as $\Delta \rightarrow 0$ the robot learns a policy that is sufficiency complex to match the given state-action pairs as precisely as possible.
At the other extreme, when $\Delta \rightarrow \infty$ the counterfactual set is equal to the entire action set $\mathcal{C}(s, a) = \mathcal{A}$. 
This in turn means that $\hat{\pi}_\theta = \pi_\theta$ in \eq{M3}, and \eq{M4} becomes the conditional entropy of $\pi_\theta(a \mid s)$.
Robots can minimize this conditional entropy with any deterministic function that outputs a consistent action at each state --- for example, simply outputting a line in \fig{method2}.
Accordingly, as $\Delta \rightarrow \infty$ the robot can fit the provided dataset with increasingly simplistic but inaccurate policies.

\subsection{Implementation} \label{sec:M3}

\begin{algorithm}[t]
    \setstretch{1.1}
    \caption{Counterfactual Behavior Cloning (Counter-BC)}
    \label{alg:1}
    \begin{algorithmic}[1]
    \State Input dataset $\mathcal{D} = \{(s_1, a_1), \ldots, (s_n, a_n)\}$
    \State Initialize control policy $\pi_\theta(a \mid s)$ with weights $\theta$
    \State Specify the radius $\Delta \geq 0$ used for sampling counterfactual actions
    \State Sample counterfactual actions $a' \in \mathcal{C}(s, a)$ for each $(s, a) \in \mathcal{D}$ \Comment \eq{M2}
    \For{epoch $\in 1, 2, \ldots, K$}
        \State $\mathcal{L}(\theta) \gets 0$ \Comment Initialize loss
        \For{$(s, a) \in \mathcal{D}$}
            \State $\log{\pi_\theta} \gets \big[\log{\pi_\theta(a' \mid s)}$ for each $a' \in \mathcal{C}(s, a)\big]$ \Comment Compute vector of length $|C(s, a)|$
            \State $\hat{\pi}_\theta \gets \text{softmax}(\log{\pi_\theta})$ \Comment \eq{M3}
            \State $\mathcal{L}(\theta) \gets \mathcal{L}(\theta) - \hat{\pi}_\theta \cdot \log{\pi_\theta}$ \Comment \eq{M4} with dot product between vectors
        \EndFor
        \State $\theta \gets \theta - \alpha \nabla_\theta \mathcal{L}(\theta)$ \Comment Update model weights to minimize loss
    \EndFor
    \State Return trained control policy $\pi_\theta(a \mid s)$
    \end{algorithmic}
\end{algorithm}

Equipped with our understanding of the loss function\textcolor{green!60!black}{,} we are now ready to present Counter-BC (see Algorithm~\ref{alg:1}).
Counter-BC is an offline imitation learning approach that inputs a dataset of state-action pairs $\mathcal{D}$ and the designer-specified hyperparameter $\Delta$.
During training\textcolor{green!60!black}{,} Counter-BC learns a control policy with parameters $\theta$ to minimize \eq{M4}: this training phase does not require any human labels or environment interactions.
Code for implementing Counter-BC is located here: \url{https://github.com/VT-Collab/Counter-BC}

In the experiments reported below we instantiate $\pi_\theta$ as a Gaussian policy with independent covariance.
This policy is constructed from a fully connected neural network with two hidden layers and ReLU($\cdot$) activation functions.
Actions are sampled from the policy using the reparameterization trick.
To train the policy network \textcolor{black}{we employ an Adam optimizer with a learning rate of $1e^{-3}$.
We selected the learning rate, batch size, and number of epochs through preliminary testing with standard behavior cloning (see our public repository for additional hyperparameter values). 
Unless otherwise stated, we normalized the actions to a unit sphere, and used $\Delta = 0.4$ in our experiments.
This value is based on the tradeoff observed in \fig{method2}.}

\textcolor{black}{
When designers are implementing Counter-BC we recommend normalizing the action space to a unit sphere, and then selecting radius $\Delta$ between $0.2$ and $0.5$. 
If the robot is learning from a naive user who often makes mistakes, the value of $\Delta$ should likely be increased towards $0.5$. 
Conversely, if the users provide high quality demonstrations, then the value of $\Delta$ should be decreased towards $0.2$.
We emphasize that these guidelines are heuristics, and the process of selecting $\Delta$ is one of the limitations of Counter-BC.
}
\textcolor{black}{We sample the counterfactual actions $a' \in \mathcal{C}(s, a)$ before the training loop (Line 4 of Algorithm~\ref{alg:1}) for computational speed; these actions do not need to be resampled or updated throughout the learning process. On average, we found that Counter-BC took roughly $5X$ longer to train than standard behavior cloning.}
\section{Simulations} \label{sec:sims}

In this section we compare Counter-BC to state-of-the-art alternatives in simulated environments.
Actual participants and synthetic users provide demonstrations within each environment, and we leverage behavior cloning variants to try and recover the intended task from these imperfect datasets.
Our simulations explore i) how different types of noise and suboptimality affect the learner's performance, ii) how the learner's behavior changes as the amount of data varies, iii) how the level of noise in the human's demonstrations impacts the learner, and finally iv) how adjusting hyperparameter $\Delta$ alters results with Counter-BC.

\p{Environments}
We selected four different environments to perform our simulations: \textit{Intersection}, \textit{Cartpole}, \textit{Car Racing}, and \textit{Robomimic}.
Images of the environments are shown in \fig{sims1}.
Overall, these environments and tasks span different levels of complexity --- from balancing a one-DoF inverted pendulum, to driving a vehicle based on RGB images, to controlling a robot arm that grasps and manipulates objects.
In order to collect real human demonstrations, we needed to choose environments where humans can actually control the ego agent (i.e., teleoperate the robot).
We also tried to pick environments that were consistent with related works, and leveraged existing datasets for learning from imperfect humans.
Below we briefly summarize each environment.

\ip{Intersection} A multi-agent driving task from \cite{mehta2025stable}.
Two cars are crossing an intersection: the ego agent (which is controlled by a simulated or real human) and another agent (which is fully automated, and updates its motion in response to the ego agent).
Both vehicles are attempting to reach their respective goals on the opposite side of the intersection while avoiding collisions with one another.
The state $s \in \mathbb{R}^4$ includes the $x$-$y$ position of both cars, and the action $a \in \mathbb{R}^2$ updates the position of the ego agent.
The performance of the robot learner is measured by its total reward across an interaction; this reward is the negative of Equation (12) in \cite{mehta2025stable}. 
\textcolor{black}{Each demonstrated trajectory contains $20$ state-action pairs.}

\ip{Cartpole} A standard environment for robot learning \cite{towers2024gymnasium}.
The state $s \in \mathbb{R}^4$ captures the pose of a cart and attached pendulum, and actions $a \in \mathbb{R}^1$ move the cart left or right along a continuous action space in order to balance the inverted pendulum.
The robot's performance is measured by the number of timesteps the agent can keep the pendulum from falling over (up to a maximum of $200$ timesteps).
Related papers on learning from noisy human demonstrations have leveraged this environment or similar variations of the inverted pendulum problem \cite{chen2021learning, sun2023offline, tangkaratt2020variational, yang2021trail, zolna2020offline}.
\textcolor{black}{On average, a human demonstration of the task contains $109$ state-action pairs.}

\ip{Car Racing} A single-agent driving along a randomly generated road \cite{towers2024gymnasium}.
\textcolor{black}{Unlike our other environments --- where the robot has direct access to the system pose --- here the robot observes $96 \times 96$ RGB images of the scene.
Specifically, the robot's state is a sequence of four frames showing the environment at timesteps $t$, $t-1$, $t-2$, and $t-3$.
The learner must map this sequence of visual inputs into continuous actions} $a \in \mathbb{R}^3$ that steer, accelerate, and brake.
Humans provide demonstrations where they try to quickly advance while keeping the car on the road; the learner's performance is measured by the distance traveled.
\textcolor{black}{For this environment we use a CNN to encode the observed images into state vectors. This CNN is trained alongside the robot's Gaussian policy.}
\textcolor{black}{On average, a human demonstration of the task contains $2225$ state-action pairs.}

\ip{Robomimic} An existing dataset for learning robot manipulation tasks from real human demonstrations \cite{mandlekar2021matters}.
The dataset includes multiple tasks: we implement the \textit{Can} task and \textit{Multi-Human} dataset.
In this task a FrankaEmika robot arm reaches to grasp a cylindrical can, carry it, and then drop it in the appropriate bin.
The system state $s \in \mathbb{R}^{23}$ includes the robot's pose, the can's pose, and the can's pose relative to the robot arm.
Actions $a \in \mathbb{R}^7$ move the robot's end-effector and actuate its gripper.
The multi-human dataset includes $300$ \textit{Can} demonstrations collected from six different humans: two ``better'' operators, two ``okay'' operators, and two ``worse'' operators.
Similar to \cite{beliaev2022imitation, kuhar2023learning, yue2024leverage}, we directly leverage this existing multi-human dataset; no additional or synthetic demonstrations are collected for this environment.
Performance is measured by the percentage of trials where the robot learner successfully completes the entire manipulation task.
\textcolor{black}{On average, a human demonstration of the task contains $209$ state-action pairs, but this varies across the human's proficiency level \cite{mandlekar2021matters}.}

\p{Human Demonstrations}
In \textit{Intersection}, \textit{Cartpole}, and \textit{Car Racing} environments we collected real and synthetic human demonstrations.
Within the synthetic demonstrations we injected different types of controlled noise or suboptimal behaviors.
A list of demonstrators is provided below:
\begin{itemize}
    \item \textbf{Real Humans.} We recruited $N=20$ in-person participants ($3$ female, average age $24 \pm 4$ years) to teleoperate the ego agent. \textcolor{black}{These participants were Virginia Tech students that provided informed written consent following university guidelines (IRB \#$23$-$1237$)}. After practicing the tasks for up to five minutes, users provided demonstrations to convey the desired behavior in each environment. \textcolor{black}{In all our experiments the participants were instructed to do their best when teaching the robot, while acknowledging that mistakes happen and the user does not always need to be perfect. Participants started with \textit{Intersection}, where they completed $50$ interactions to provide ($1000$ state-action pairs). Next, participants teleoperated the simulated \textit{Cartpole} for $1000$ timesteps, and finally participants drove \textit{Car Racing} one complete lap around its track. No participant data was excluded. The state-action pairs from all participants were aggregated into a single dataset for each environment, and then the algorithms listed below were trained on a randomly sampled subset of state-action pairs sampled from this aggregated dataset.}
    \textcolor{black}{Participants were compensated at a rate of \$20USD/hour for their time. The total study duration was approximately 45--60 minutes per participant, including a consent and briefing period ($\approx$5 minutes), practice trials in each simulated environment ($\approx$15 minutes total), data collection across the three simulated environments ($\approx$15 minutes), and the air hockey robot experiment ($\approx$10 minutes including setup and recorded demonstrations, see Section~\ref{sec:user-study}).}
    \item \textbf{Uniform.} Synthetic humans demonstrated the task by sampling actions according to \eq{P1}, where $a^*$ is the optimal action that maximizes task performance and $P$ is a uniform distribution: $\epsilon \sim \mathcal{U}(-\sigma, \sigma)$. Intuitively, these simulated human teachers had uniform noise added to their demonstrations.
    \item \textbf{Gaussian.} Synthetic humans sampled actions according to \eq{P1}, where $a^*$ is the optimal action and $P$ is an unbiased Gaussian distribution: $\epsilon \sim \mathcal{N}(0, \sigma^2)$.
    \item \textcolor{black}{\textbf{$\epsilon$-Greedy.}} With probability $1 - \sigma$ the demonstrator provides the optimal action so that $a = a^*$. Otherwise the synthetic human injects uniform noise: $\epsilon \sim \mathcal{U}(-\sigma, \sigma)$.
\end{itemize}
We emphasize that collecting actual human data is critical since the purpose of our approach is to learn from humans; \textit{synthetic data may never capture the types of mistakes that real users make}.
At the same time, testing with synthetic data is theoretically useful because it provides evidence about which noise distributions are most advantageous or challenging for each method.
\textcolor{black}{
We performed offline testing to qualitatively compare the human datasets with our synthetic options.
In \textit{Intersection} the human-provided demonstrations have an average reward of $12.3$, while the synthetic demonstrations have an average reward of $13.1$.
In \textit{Cartpole} the average human and synthetic reward is $30.3$ and $26.6$, respectively.}
\textcolor{black}{These similar average rewards suggest that the synthetic demonstrations are not entirely dissimilar from the human's behavior in terms of task performance. However, we do not claim that this comparison establishes equivalence of noise structure. Testing with real human data remains essential, and the synthetic experiments are intended to provide controlled insights into how different noise types affect each algorithm.}

\p{Learning Algorithms}
Given a dataset $\mathcal{D}$ of state-action pairs provided by the real or synthetic human, the robot used offline imitation learning to train a policy $\pi_\theta$.
We compared policies learned with \textbf{Counter-BC} (Algorithm~\ref{alg:1}) to three state-of-the-art baselines for learning from noisy humans.
\begin{itemize}
    \item \textbf{BC} \cite{ke2021imitation}: Standard behavior cloning. The robot trains a Gaussian policy to try and exactly match the actions demonstrated by the human teacher.
    \item \textcolor{black}{\textbf{Weighted-BC}: A variant of behavior cloning where the robot learns the relative expertise of different state-action pairs in the dataset. Our approach here is based on ILEED from \cite{beliaev2022imitation}. Specifically, we introduce a network $\rho_\phi(s) \rightarrow[0, 1]$ that inputs the state and outputs a normalized score. This score estimates the expertise level of the demonstrations, i.e., as $\rho_\phi(s) \rightarrow 1$ the robot is confident the actions shown at $s$ precisely demonstrate the task. The learner then sets $R(s, a) = \rho_\phi(s)$, and simultaneously trains both $\rho_\phi$ and $\pi_\theta$ using the loss from \eq{M1} with no counterfactual actions. In practice, Weighted-BC enables the learner to distinguish between states where the human teachers provide consistent, expert actions, and states where the teachers input noisy, suboptimal actions.}
    \item \textcolor{black}{\textbf{BC-RNN} \cite{mandlekar2021matters}: A version of behavior cloning where the robot reasons over a history of recent states. We combine our standard \textbf{BC} approach with a Gated Recurrent Unit (GRU). This GRU inputs a sequence of $5$ states, and outputs a latent state representation based on the state history. The Gaussian policy then reasons over the latent state to select actions that match the behaviors of the human teacher. Our intuition is that this history-based approach may be able to disambiguate between local noise and underlying action patterns.}
    \item \textbf{BCND} \cite{sasaki2020behavioral}: A behavior cloning variant that shifts the learned policy towards the mode of the distribution. For this approach we set $R(s, a) = \pi_{prev}(a \mid s)$, the policy trained at the previous iteration. In practice \textbf{\textcolor{black}{BCND}} learns to emulate the human's actions at states where the demonstrations are consistent, and ignores the human's actions with high variability.
\end{itemize}
Note that we did not include offline reinforcement learning algorithms since the robot learner does not have access to the task reward, and thus these methods cannot be applied to our setting.
In our implementation Counter-BC and the baselines are instantiated as Gaussian policies.
The learning rate, batch size, and number of epochs are held constant across all methods to mitigate against biased results. 
\textcolor{black}{We note that ILEED \cite{beliaev2022imitation} was originally designed for settings where there are multiple human teachers, and the robot knows which demonstrations belong to which teacher.
The \textbf{Weighted-BC} baseline --- which functions as a simplified version of ILEED --- may therefore perform poorly because we do not assume any labels on the dataset, and so \textbf{Weighted-BC} cannot connect individual datapoints with any specific human teacher.}
Additional training details can be found in our public code repository: \url{https://github.com/VT-Collab/Counter-BC}

\begin{figure*}[t]
	\begin{center}
        \includegraphics[width=\linewidth]{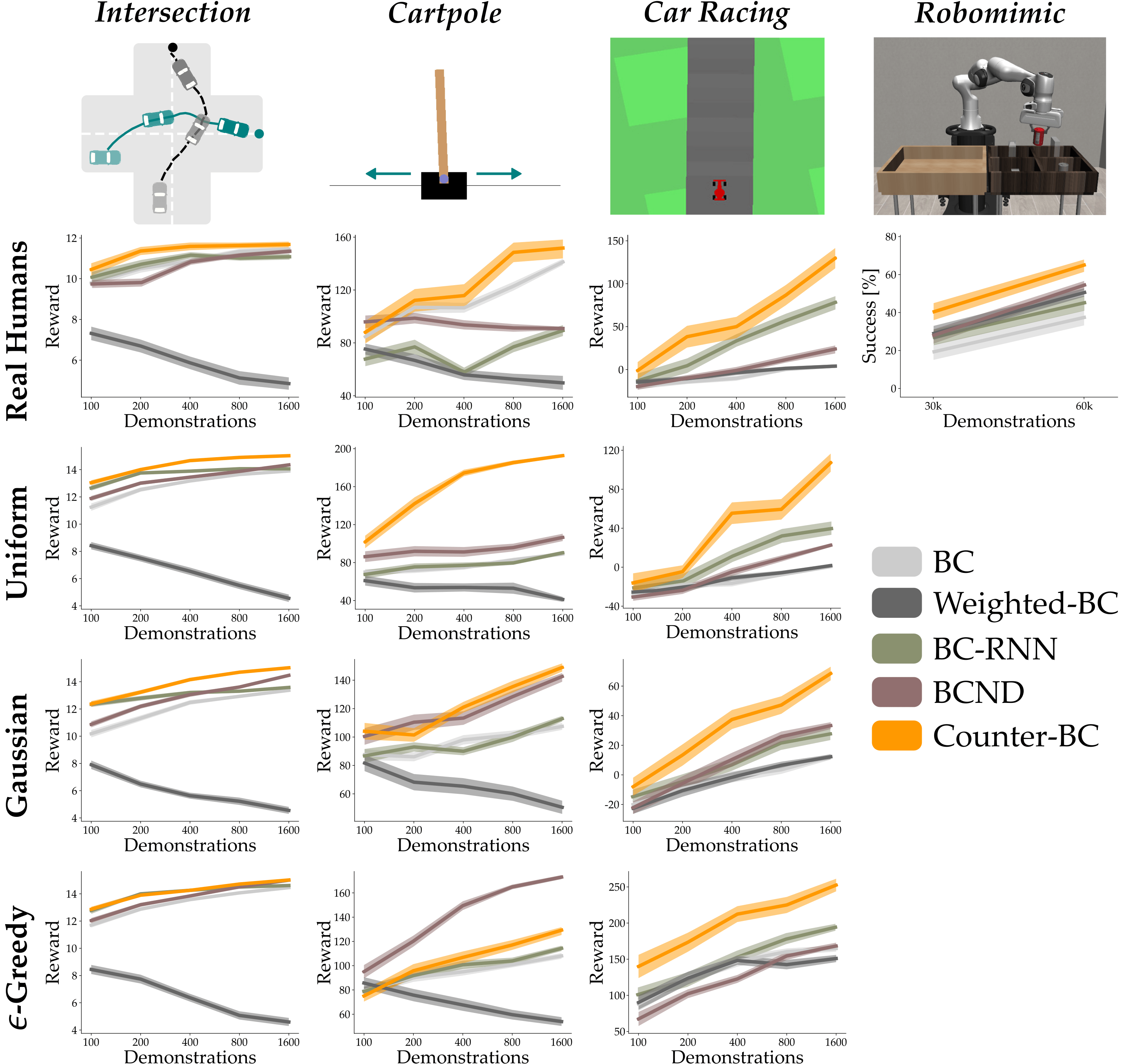}
		\caption{Learning from real and simulated humans (see Sections~\ref{sec:S1} and \ref{sec:S2}). Every column shows a different environment, and the rows correspond to demonstrations from real humans, or synthetic operators \textcolor{black}{\sout{with}} uniform, Gaussian, and \textcolor{black}{$\epsilon$-Greedy}. \textcolor{black}{Here ``demonstrations'' refers to the number of state-action pairs in the training dataset}. Plots illustrate the performance of the policy learned by each algorithm as a function of the number of demonstrated state-action pairs (higher reward or success is better). For \textit{Robomimic} we tested the \textit{Can} task with standardized demonstrations from the Multi-Human dataset \cite{mandlekar2021matters}. Results are averaged across $50$ runs \textcolor{black}{with different seeds}; shaded regions show standard error. \textcolor{black}{A repeated measures ANOVA shows that algorithm type had a statistically significant effect on reward across all environments and human teachers. Post hoc tests confirm that \textbf{Counter-BC} led to higher rewards on average as compared to \textbf{BC} ($p<.01$), \textbf{Weighted-BC} ($p<0.001$), \textbf{BC-RNN} ($p<.01$) and \textbf{BCND} ($p<.01$).}}
		\label{fig:sims1}
	\end{center}
        \vspace{-1em}
\end{figure*}

\subsection{How Does Counter-BC Perform with Different Types of Imperfect Demonstrations?} \label{sec:S1}

The results of our first simulation are shown in \fig{sims1}.
Each row of this figure corresponds to a different type of human demonstrator: actual participants, and then synthetic teachers with uniform, Gaussian, or \textcolor{black}{$\epsilon$-Greedy}.
For the \textit{Robomimic} environment we only collect data from real humans by leveraging the standardized dataset in \cite{mandlekar2021matters}.
In general, we observe that \textbf{Counter-BC} learns more proficient policies than the baselines.
We particularly highlight the performance of \textbf{Counter-BC} in the \textit{Car Racing} environment --- where the control policies are based on image observations, and not direct state measurements --- as well as the performance across environments with real human demonstrations.
By contrast, the one type of imperfect demonstration where \textbf{Counter-BC} does not consistently reach the highest reward is \textcolor{black}{$\epsilon$-Greedy}; here \textbf{\textcolor{black}{BCND}} occasionally outperforms \textbf{Counter-BC}.
This result makes sense because \textbf{\textcolor{black}{BCND}} is designed to work with \textcolor{black}{$\epsilon$-Greedy}.
More specifically, \textbf{\textcolor{black}{BCND}} trains the policy so that it is better aligned with the human teacher's most common actions --- and when our simulated humans \textcolor{black}{are $\epsilon$-Greedy}, their most common action is the optimal action $a^*$.
For all other types of imperfect demonstrations we find \textbf{Counter-BC} leads to improved performance.
Taken together, this suggests that \textbf{Counter-BC} learns the intended task despite controlled noise or uncontrolled errors in the human's demonstrations.

\subsection{How Does Counter-BC Perform with an Increasing Number of Demonstrations?} \label{sec:S2}

Within \fig{sims1} we also measure how the performance of each algorithm changes with varying amounts of human data.
\textcolor{black}{The number of demonstrated state-action pairs is shown along the $x$-axis of each individual plot.}
Intuitively, we would expect the system to perform better when given more examples: the more times the human shows the robot how to act, the more proficiently it should behave during rollouts.
Our results affirm this intuition.
Additionally, we see that \textbf{Counter-BC} outperforms the alternatives across different levels of demonstrations.
\textcolor{black}{This suggests that \textbf{Counter-BC} is not limited to settings with a specific number of demonstrations, but can be used across varying dataset sizes. Additional information about the percent improvement between \textbf{Counter-BC} and the baselines can be found in the Appendix (Section~\ref{sec:appendix}).}

\begin{figure*}[t]
	\begin{center}
        \includegraphics[width=\linewidth]{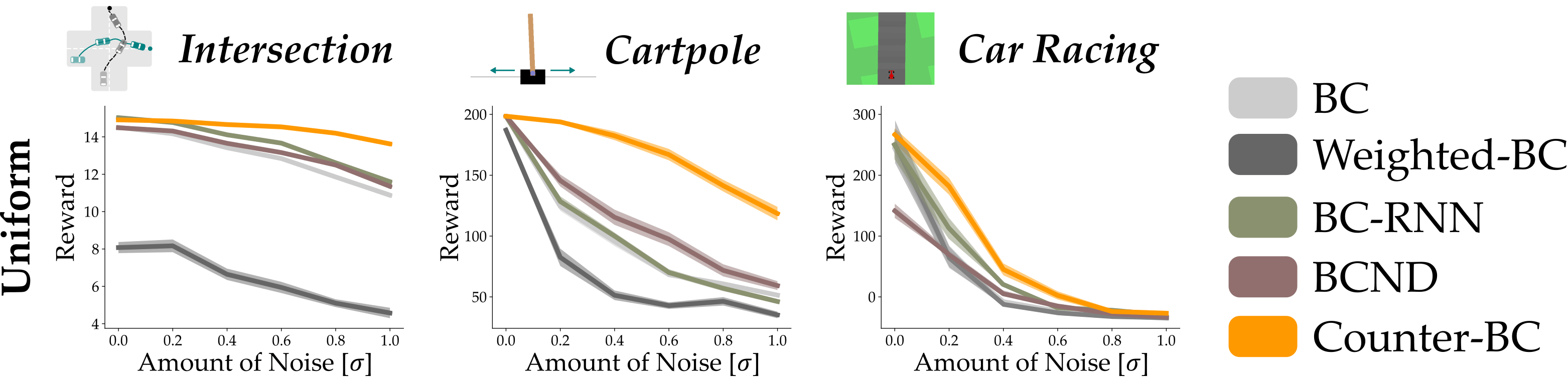}
		\caption{Learning from increasingly noisy demonstrations (see Section~\ref{sec:S3}). To regulate the amount of noise, we simulated a human teacher. This teacher selected actions $a = a^* + \epsilon$, where $a^*$ is the optimal action and $\epsilon \sim \mathcal{U}(-\sigma, \sigma)$ is uniform noise. Increasing $\sigma$ led to more noisy, imperfect, and suboptimal human teaching. The policies learned by each algorithm degrade as $\sigma$ increases, but \textbf{Counter-BC} is more robust than the baselines. Results are averaged across $50$ runs with demonstrations containing $400$ state-action pairs. \textcolor{black}{We conducted a repeated measures ANOVA using this data. Post hoc tests show that \textbf{Counter-BC} led to higher average rewards than \textbf{BC} ($p<.01$), \textbf{Weighted-BC} ($p<0.01$), \textbf{BC-RNN} ($p<.01$) and \textbf{BCND} ($p<.01$).}}
		\label{fig:sims2}
	\end{center}
        \vspace{-1em}
\end{figure*}

\subsection{How Does Counter-BC Perform with Varying Levels of Noise?} \label{sec:S3}

So far we have varied the type of human teacher and the amount of data; next, we adjust the level of noise within the human's examples.
\fig{sims2} corresponds to a setting where the robot is learning from increasingly suboptimal and imperfect demonstrations.
When $\sigma = 0$ the synthetic human teacher perfectly demonstrates the task with optimal actions $a^*$, and as $\sigma$ increases the human deviates from the optimal actions with uniform noise $\epsilon \sim \mathcal{U}(-\sigma, \sigma)$.
Within the \textit{Intersection}, \textit{Cartpole}, and \textit{Car Racing} environments the maximum magnitude of an action dimension is $1$; hence, $\sigma = 1$ means that the size of the noise matches or exceeds the size of the optimal actions.
We find that --- relative to the baselines --- \textbf{Counter-BC} is increasingly robust as $\sigma$ grows and the synthetic human's demonstrations become more noisy and suboptimal.
When the human's demonstrations have no noise ($\sigma \rightarrow 0$) then each method is roughly equivalent.
But as the size of the noise increases ($\sigma \rightarrow 1$) we see the advantages of \textbf{Counter-BC} in reasoning over counterfactual actions, searching for simple policy explanations, and extrapolating the underlying task.
\textcolor{black}{We note that for \textit{Car Racing} at high noise levels ($\sigma \rightarrow 1$), all methods approach chance performance. This environment is particularly sensitive to noise because: (i) the policy maps high-dimensional image observations to actions, requiring a reliable association between visual features and correct steering/braking commands; and (ii) the training datasets used in this experiment ($400$ state-action pairs) represent only a fraction of a full lap, making it difficult to learn a globally consistent control policy under heavy noise. At lower noise levels and with more data, \textbf{Counter-BC} continues to outperform the baselines in this environment (see \fig{sims1} and \fig{sims2}).}

\begin{figure*}[t]
	\begin{center}
        \includegraphics[width=\linewidth]{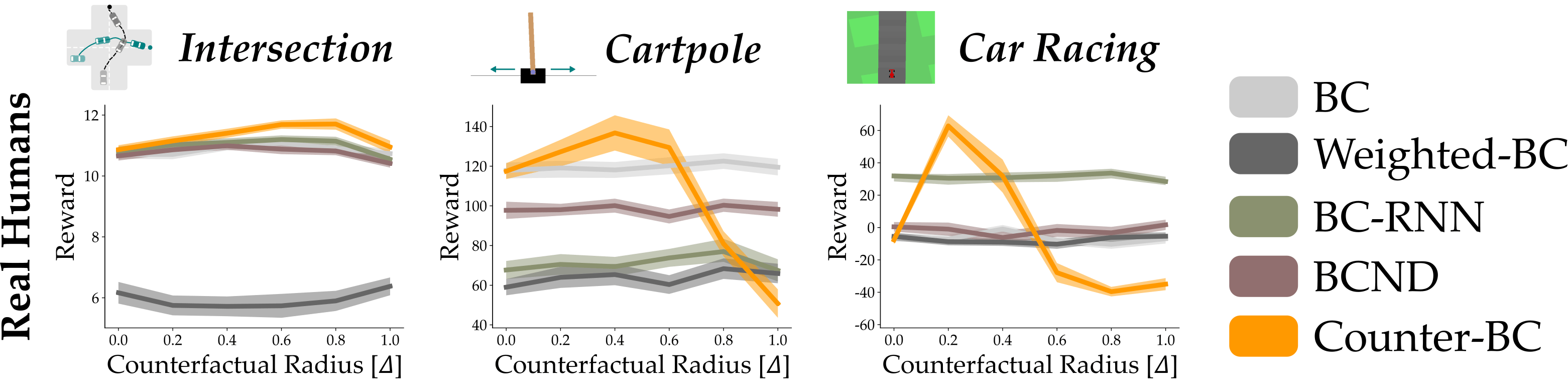}
		\caption{Learning from demonstrations while varying hyperparameter $\Delta$ (see Section~\ref{sec:S4}). \textcolor{black}{Note that the hyperparameter $\Delta$ only applies to \textbf{Counter-BC}, but the performance of the baselines slightly varies across the $x$-axis because the training dataset and testing start states are sampled randomly at each run.} Consistent with our theoretical analysis from Section~\ref{sec:M2}, when $\Delta = 0$ then \textbf{Counter-BC} is equivalent to \textbf{BC}. Increasing $\Delta$ enlarges the counterfactual sets and leads to simpler explanations of the human's demonstrations. Once $\Delta$ is increased too much, the robot begins to oversimplify the underlying human policy, resulting in degraded performance. Shaded regions show the standard error across $50$ end-to-end runs with demonstrations that have $400$ state-action pairs. These demonstrations were collected from $N=20$ in-person participants.}
		\label{fig:sims3}
	\end{center}
        \vspace{-1em}
\end{figure*}

\subsection{How Does Adjusting Hyperparameter $\Delta$ Affect Counter-BC?} \label{sec:S4}

We conclude our simulations by testing the effect of hyperparameter $\Delta$ on \textbf{Counter-BC}.
In Section~\ref{sec:M2} and \fig{method2} we theoretically showed that $\Delta \rightarrow 0$ causes the robot to exactly imitate all of the human's actions, reducing \textbf{Counter-BC} to standard \textbf{BC}.
Conversely, as $\Delta$ increases the robot should learn simpler functions by imitating actions close to (but not the same as) the behaviors actually demonstrated by the human teacher, resulting in learning capabilities beyond the current state-of-the-art.
In \fig{sims3} we put this theory to the test with real human data.
For datasets from $N=20$ participants in \textit{Intersection}, \textit{Cartpole}, and \textit{Car Racing}, we learn robot policies while adjusting the value of $\Delta$.
As expected, when $\Delta = 0$ the performance of \textbf{Counter-BC} equals the behavior of \textbf{BC}.
Increasing $\Delta$ enables \textbf{Counter-BC} to expand the human's demonstrations and extrapolate underlying functions, improving performance of the learned policy.
However, when $\Delta$ is increased too far the robot eventually recovers policies that are simpler than what the human intended (i.e., missing nuances in the correct behavior), harming the performance of the autonomous robot.
This suggests both a trade-off and guidelines for implementation.
We recommend that designers start with low values of $\Delta$, and gradually increase that hyperparameter to improve performance.
We also recognize that one downside of \textbf{Counter-BC} is that --- if designers initialize with values of $\Delta$ that significantly exceed the noise in the teacher's demonstrations --- the performance of our algorithm may actually fall below the \textbf{BC} baseline.
\section{User Study} \label{sec:user-study}

Our previous experiments focused on simulated environments with real and synthetic human teachers.
In our final study, we now test Counter-BC on a real-world robot arm (Franka FR3) with demonstrations provided by $N=20$ in-person participants.
The task matches our running example from \fig{front}: human teachers teleoperate the robot arm to repeatedly hit an air hockey puck across a table.
Based on the participants' demonstrations, the robot must learn to align its paddle with the oncoming puck, and then move forward to strike that puck with the correct timing.
See examples of this game in our supplemental video:
\url{https://youtu.be/XaeOZWhTt68}

\p{\textcolor{black}{Data Collection}}
\textcolor{black}{Our experiment has two phases: data collection and evaluation.}
During \textit{data collection} each individual participant sits next to the robot and uses a joystick to teleoperate the arm in real-time.
The arm's end-effector is constrained to move along the planar surface of an air hockey table.
A red puck glides along this table; participants try to control the robot to hit that puck so it bounces off the opposite side and returns towards the robot arm.
Based on the speed and location where participants hit the puck, the puck may move at various angles and velocities (e.g., colliding with the table's sides).
Hence, each iteration naturally forces users to vary the way they move the robot arm.
\textcolor{black}{In total, $N=20$ users provided demonstrations for the air hockey task; 
these were the same users that taught the simulated robots in Section~\ref{sec:sims}.
Participants practiced teleoperating the robot arm with a joystick for $2$ to $5$ minutes.
Users then controlled the robot to play air hockey for $2$ minutes of recorded demonstrations.
This corresponds to roughly $2400$ state-action pairs per user.}
These participants are not perfect teachers --- their motions include examples where they miss the puck or accidentally hit it off-center.

\begin{figure*}[t]
	\begin{center}
        \includegraphics[width=\linewidth]{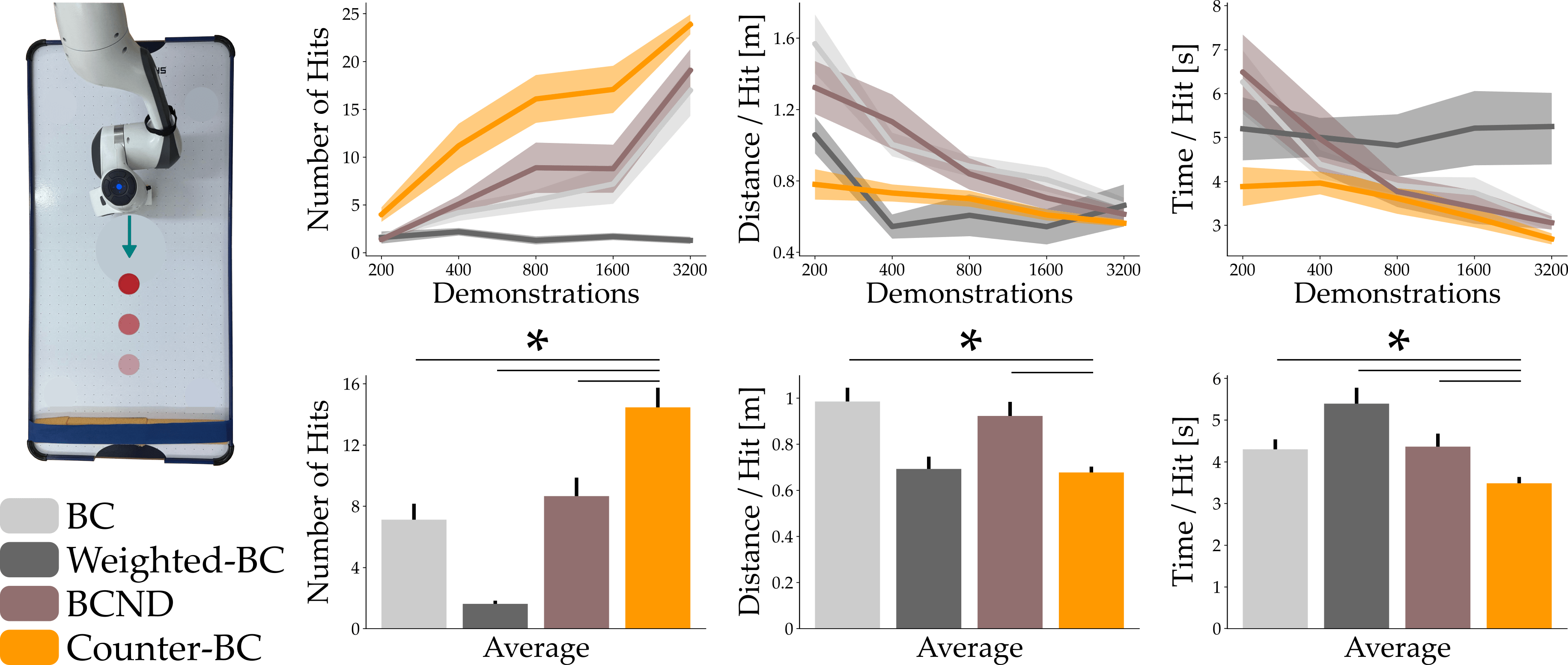}
		\caption{Results from our air hockey experiment in Section~\ref{sec:user-study}. In-person participants provided demonstrations where they teleoperated the robot arm to repeatedly hit a hockey puck (also see \fig{front}). \textcolor{black}{The top row plots the performance of the robot's learned policy when trained on an increasing number of state-action pairs (i.e., ``demonstrations'')}; the bottom row shows the average performance across all dataset sizes. \textit{Number of Hits} is the number of times the robot autonomously hit the puck (higher is better). \textit{Distance per Hit} is path length of the robot's end-effector divided by the number of hits (lower is better), and \textit{Time per Hit} is the total interaction time divided by the number of hits (lower is better). Our results suggest that robots which use \textbf{Counter-BC} to learn from human demonstrations outperform other offline imitation learning approaches that account for imperfect teaching. Shaded regions and error bars show the standard error, and an $*$ denotes statistical significance ($p < .05$).}
		\label{fig:hockey}
	\end{center}
        \vspace{-1em}
\end{figure*}

Throughout data collection the robot records the position of the puck with an overhead camera (see \fig{hockey}, Left).
The robot also saves its end-effector locations and the user's commanded actions.
Overall, the robot's state $s \in \mathbb{R}^6$ consists of the $x$-$y$ position of its end-effector, the $x$-$y$ position of the puck at the current timestep, and the puck's $x$-$y$ position at the previous timestep.
Actions $a \in \mathbb{R}^2$ are end-effector velocities along the surface of the table normalized between $0$ and $1$.
Each timestep of data collection yields a new $(s, a)$ pair for our aggregated dataset $\mathcal{D}$.

\p{Independent Variables}
\textcolor{black}{After collecting dataset $\mathcal{D}$ we next move to the \textit{evaluation} phase. 
During evaluation the robot trains its policy using subsets of the aggregated demonstrations, and then rolls out this learned policy to autonomously play air hockey.}
We modulate two variables: the \textit{algorithm} the robot uses to train its control policy, and the \textit{amount of data} leveraged by that algorithm.
We compare the same offline imitation learning methods as in Section~\ref{sec:sims}, with the exception of BC-RNN.
These include our proposed \textbf{Counter-BC} approach, as well as state-of-the-art baselines \textbf{BC} \cite{ke2021imitation}, \textbf{\textcolor{black}{Weighted-BC}} \cite{beliaev2022imitation}, and \textbf{\textcolor{black}{BCND}} \cite{sasaki2020behavioral}.
Each algorithm learns from the same human demonstrations.
We sample these demonstrations from the aggregated dataset $\mathcal{D}$; specifically, we measure the robot's performance when trained on $200$, $400$, $800$, $1600$, and $3200$ state-action pairs.
\textcolor{black}{For reference, a dataset size of $200$ corresponds to $10$ seconds of human demonstration, and a dataset size of $3200$ corresponds to $160$ seconds of human demonstration.
We chose these dataset sizes to be roughly consistent with the simulations in Section~\ref{sec:sims} and \fig{sims1}; in our preliminary testing, $100$ state-action pairs was not sufficient for any method to learn a meaningful air hockey policy.}

During evaluation we perform $10$ end-to-end runs for each dataset size.
\textcolor{black}{We emphasize that the robot policies are re-trained from scratch at each run, so the results presented below show performance across $50$ separate rounds of training and testing per algorithm.}
In what follows we describe a single one of these runs.
To evaluate the performance with $k$ datapoints, we first sample $k$ state-action pairs from dataset $\mathcal{D}$ and then train all four algorithms with those same datapoints.
For each method (\textbf{BC}, \textbf{\textcolor{black}{Weighted-BC}}, \textbf{\textcolor{black}{BCND}}, \textbf{Counter-BC}) we place the hockey puck directly in front of the robot's end-effector and start executing the learned policy.

\p{Dependent Measures}
Ideally, the robot will learn to repeatedly hit the puck while moving quickly and efficiently.
We therefore count the \textit{Number of Hits}, i.e., the number of times the robot uses its end-effector to push the puck across the table.
To assess efficiency, we record the \textit{Distance per Hit}.
This corresponds to the distance the robot's end-effector travels during an interaction (in meters) divided by the number of hits; if the robot never hits the puck, then this is just the total distance traveled.
Similarly, the \textit{Time per Hit} is the total interaction time (in seconds) divided by the number of hits.
As before, if the robot never hits the puck, \textit{Time per Hit} becomes the total interaction time.
A proficient robot should repeatedly hit the puck (maximizing \textit{Number of Hits}) while minimizing the distance traveled (\textit{Distance per Hit}) and the time between hits (\textit{Time per Hit}).

\p{Hypotheses} We hypothesized that:
\begin{quote}
\textit{Robots that use \textbf{Counter-BC} to learn from human demonstrations will extract more proficient control policies than the baselines.}
\end{quote}

\p{Results}
Our experimental results are summarized in \fig{hockey}.
In the top row we plot our dependent measures as a function of the number of demonstrations; the bottom row displays the average outcomes across all $50$ trials for each algorithm.
As a reminder, for the \textit{Number of Hits} \textbf{higher} values indicate better task performance.
By contrast, for the \textit{Distance per Hit} and the \textit{Time per Hit} \textbf{lower} values mean the robot is playing the game more quickly and efficiently.

To capture overall trends in the learned policies we will analyze the average results.
Repeated measures ANOVAs with Greenhouse-Geisser corrections determine that algorithm type has a statistically significant effect on the \textit{Number of Hits} ($F(2.74, 134.24) = 39.8$, $p < .001$), the \textit{Distance per Hit} ($F(2.42, 118.78) = 12.6$, $p < .001$), and the \textit{Time per Hit} ($F(1.89, 92.48) = 8.3$, $p < .001$).
To better understand these differences, we next conduct post hoc tests with a Bonferroni adjustment.
Here we see that \textbf{Counter-BC} averages a higher \textit{Number of Hits} across all demonstrations as compared to \textbf{BC} ($p<.001$), \textbf{\textcolor{black}{Weighted-BC}} ($p<.001$), and \textbf{\textcolor{black}{BCND}} ($p<.001$).
Similarly, the \textit{Time per Hit} for \textbf{Counter-BC} is lower than \textbf{BC} ($p<.01$), \textbf{\textcolor{black}{Weighted-BC}} ($p<.001$), and \textbf{\textcolor{black}{BCND}} ($p<.05$).
For the \textit{Distance per Hit} we gather mixed findings: \textbf{Counter-BC} is again more performant than \textbf{BC} ($p <.001$) and \textbf{\textcolor{black}{BCND}} ($p < .01$), but comparisons with \textbf{\textcolor{black}{Weighted-BC}} are not statistically significant ($p = 1.0$).
We can explain this result by watching the learned behavior for \textbf{\textcolor{black}{Weighted-BC}}.
\textcolor{black}{\textit{Distance per Hit} reflects the distance the robot traveled during each interaction. In our tests, robots trained with \textbf{Weighted-BC} remained close to their initial position, often missing the puck (as reflected by the low \textit{Number of Hits}) or hitting that puck at a low velocity (as reflected by the high \textit{Time per Hit}). Accordingly, while the \textbf{Weighted-BC} robots learned concise motions similar to \textbf{Counter-BC}, these motions were not effective at completing the overall task.}

\p{\textcolor{black}{Discussion}}
In summary, our results from \fig{hockey} support our hypothesis and suggest that \textbf{Counter-BC} is an effective algorithm for learning from real human data on physical robot arms.
Across each axis of evaluation (\textit{Number of Hits}, \textit{Distance per Hit}, and \textit{Time per Hit}), we find that \textbf{Counter-BC} outperforms or matches the baselines, and these trends hold regardless of the number of state-action pairs leveraged to train the robot learner.

\textcolor{black}{
More generally, our experiments in Section~\ref{sec:sims} and Section~\ref{sec:user-study} included learning tasks from images, learning tasks with high-dimensional states, and learning tasks on real-world robot arms.
The combined results indicate that Counter-BC often learns more proficient policies than state-of-the-art baselines.
In particular, we found that Counter-BC is robust to different types of human data with varying levels of noise, and that Counter-BC is also more performant when trained on real human demonstrations.
One of the strengths of Counter-BC is its relatively simple implementation (see \url{https://github.com/VT-Collab/Counter-BC}), and its theoretical connection to other behavior cloning variants.
We believe that Counter-BC is well suited for scenarios where the robot is given a single, aggregated dataset: this dataset could be provided by one human teacher, or the result of multiple different users interacting with the system.
We also highlight that Counter-BC can be used in combination with other approaches (e.g., diffusion policies) by simply adding a counterfactual within those approaches and then minimizing the entropy of the learned policy.
On the other hand, Counter-BC may struggle with tasks that require multimodal policies; see our Limitations below.
}

\section{Conclusion} \label{sec:conclusion}

\textcolor{black}{Existing imitation learning algorithms seek to emulate the ``optimal'' actions of the human teacher. 
When the human's teaching is of varying quality, these approaches attempt to separate ``optimal'' and ``noisy'' actions, and then train a policy to exactly match these optimal behaviors.}
By contrast, this work introduced a different perspective.
Our core hypothesis was that all of the human's data is based on the underlying task they are trying to convey.
Suboptimality inherent to human teaching adds unintended complexity to the dataset, but we can develop learning algorithms to work around this noise and recover the underlying function.
In doing so, the robot is no longer constrained to imitate the human's \textit{exact} datapoints; instead, the robot can now propose and imitate \textit{counterfactual} actions to \textcolor{black}{extrapolate} what the human meant.
We formalized this concept by deriving Counter-BC, an offline imitation learning algorithm that accounts for imperfect human teachers.
We proved that the loss function used to train Counter-BC generalized prior works; put another way, current methods can be viewed as instances of our overarching approach.
In practice, by optimizing this generalized loss Counter-BC trains the policy to minimize entropy over actions (i.e., reaching a \textcolor{black}{confident prediction}) while ensuring that the actions output by the policy are close to the actions demonstrated by the human (i.e., remaining within the counterfactual set).
Our analysis indicated that Counter-BC can extract the desired policy from imperfect data, multiple users, and teachers of varying skill.
To support these findings, we tested Counter-BC across several environments with real human teachers, synthetic demonstrations, and standardized datasets.
Overall, our theoretical and empirical contributions are a step towards robots that learn what their human teachers actually meant, not just what the human teacher showed.

\p{Limitations}
One advantage and disadvantage of Counter-BC is the hyperparameter $\Delta \geq 0$ in Algorithm~\ref{alg:1}.
This hyperparameter specifies the size of the counterfactual set; intuitively, increasing $\Delta$ means the robot can extrapolate more from the human's data, but the resulting policy may contain greater deviations from the human's demonstrations.
The advantage of $\Delta$ is that designers can tune this parameter along a continuous spectrum so that Counter-BC works with expert teachers (low values of $\Delta$) and inexperienced teachers (high values of $\Delta$).
The disadvantage is that --- if designers choose too large a $\Delta$ --- Counter-BC may oversimplify the policy and miss important nuances in the task.
Within our manuscript we tried to mitigate this potential issue by providing recommended values and by testing with a range of $\Delta$.
In future work, we would like to automate the selection of $\Delta$ based on the offline dataset of human demonstrations.

\textcolor{black}{A second limitation of Counter-BC is its approach to multimodal human behaviors. For some tasks the learned policy should output multiple, distinct actions. Imagine a mobile robot driving towards an obstacle: ideally, the robot has learned to pass this obstacle both on the right (one mode) and on the left (a second mode). Counter-BC inherently struggles with these multimodal scenarios because it is trained to minimize action entropy within its counterfactual set, thereby penalizing any policy which tries to match more diverse human demonstrations. Instead of learning to pass on both sides of the obstacle, for instance, the policy trained with Counter-BC will likely collapse to just one of these demonstrated modes (e.g., always passing on the right). Our experimental results suggest that Counter-BC can handle some of the action diversity that is inherent to human demonstrations (e.g., our participants in Section~\ref{sec:user-study} teleoperated the robot with different timings, speeds, and movement patterns). However, we do not think Counter-BC is well suited for tasks which require multimodal policies. We plan to explore this limitation in future work.}

\textcolor{black}{A more severe form of the multimodal limitation arises when heterogeneous demonstrators have genuinely different goals (rather than different styles for achieving the same goal). Here Counter-BC would likely collapse to one task rather than representing both. Resolving this limitation likely requires first segmenting the dataset by demonstrator intent --- for example, conditioning on task --- before applying Counter-BC with the relevant state conditioning.}


\bibliographystyle{ACM-Reference-Format}
\bibliography{bibtex}


\section{Appendix} \label{sec:appendix}



\begin{table}[h]
  \centering
  \caption{Counter-BC relative improvement (\%) over each baseline across tasks. See \fig{sims1}.}
  \label{tab:counterbc-wide}
  \begin{tabular}{lrrrr}
    \toprule
    \textbf{Baseline} & \textbf{Interaction} & \textbf{Cartpole} & \textbf{Car Racing} & \textbf{Robomimic} \\
    \midrule
    BC           & 10\%   & 39\%   & 2{,}370\% & 21\% \\
    Weighted-BC  & 144\%  & 129\%  & 1{,}179\% & 13\% \\
    BC-RNN       & 8\%    & 51\%   & 276\%     & 15\% \\
    BCND         & 9\%    & 38\%   & 563\%     & 11\% \\
    \bottomrule
  \end{tabular}
\end{table}



\end{document}